\documentclass[10pt,twocolumn,letterpaper]{article}

\usepackage{iccv}
\usepackage{times}
\usepackage{epsfig}
\usepackage{graphicx}
\usepackage{amsmath}
\usepackage{amssymb}
\usepackage{booktabs}
\usepackage{multirow}
\usepackage{caption}
\usepackage{graphbox}
\usepackage[accsupp]{axessibility}
\pdfoutput=1

\usepackage{arydshln}
\usepackage{pifont}
\usepackage{stmaryrd}

\usepackage[pagebackref=true,breaklinks=true,letterpaper=true,colorlinks,bookmarks=false]{hyperref}

\iccvfinalcopy 

\def\iccvPaperID{1998} 

\ificcvfinal\fi

\newcommand*\samethanks[1][\value{footnote}]{\footnotemark[#1]}
\newcommand\blfootnote[1]{%
  \begingroup
  \renewcommand\thefootnote{}\footnote{#1}%
  \addtocounter{footnote}{-1}%
  \endgroup
}

\begin{document}

\title{Image Shape Manipulation from a Single Augmented Training Sample}

\author {Yael Vinker\thanks{Equal contribution} \qquad Eliahu Horwitz\samethanks \qquad Nir Zabari \qquad Yedid Hoshen\\
School of Computer Science and Engineering \\ The Hebrew University of Jerusalem, Israel\\
\small\url{http://www.vision.huji.ac.il/deepsim/}\\
{\tt\small \{yael.vinker, eliahu.horwitz, nir.zabari, yedid.hoshen\}@mail.huji.ac.il}\\
}

\twocolumn[{

	\maketitle
	\vspace{-3em}
	\renewcommand\twocolumn[1][]{#1}
	\begin{center}
		\centering
		\includegraphics[width=1\textwidth]{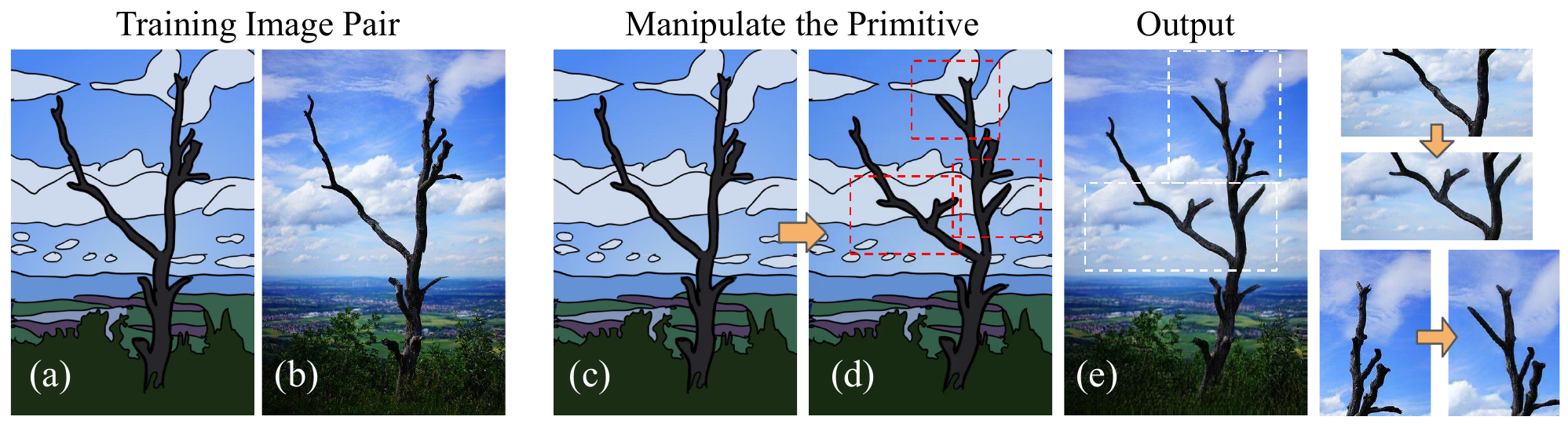}
		\vspace{-2em}
		\captionof{figure}{\textit{Image manipulation learned from a single training pair.} 
		Given a single real image (b) and a corresponding primitive representation (a), our model learns to map between the primitive (a) to the target image (b). At inference, the original primitive (c) is manipulated by the user, the changes are highlighted in red (d). Then, the manipulated primitive is passed through the network which outputs a corresponding manipulated image (e) in the real image domain. On the right, we can see that the manipulation was performed successfully, while preserving the internal statistics of the source image.}
		\label{fig:main_im}
	\end{center}
}]

\ificcvfinal\thispagestyle{empty}\fi
\begin{abstract}
In this paper, we present DeepSIM, a generative model for conditional image manipulation based on a single image. 
We find that extensive augmentation is key for enabling single image training, and incorporate the use of thin-plate-spline (TPS) as an effective augmentation.
Our network learns to map between a primitive representation of the image to the image itself.
The choice of a primitive representation has an impact on the ease and expressiveness of the manipulations and can be automatic (e.g. edges), manual (e.g. segmentation) or hybrid such as edges on top of segmentations.
At manipulation time, our generator allows for making complex image changes by modifying the primitive input representation and mapping it through the network.
Our method is shown to achieve remarkable performance on image manipulation tasks.
\end{abstract}

\begin{figure*}[t]

\begin{center}
\includegraphics[width=0.9\linewidth]{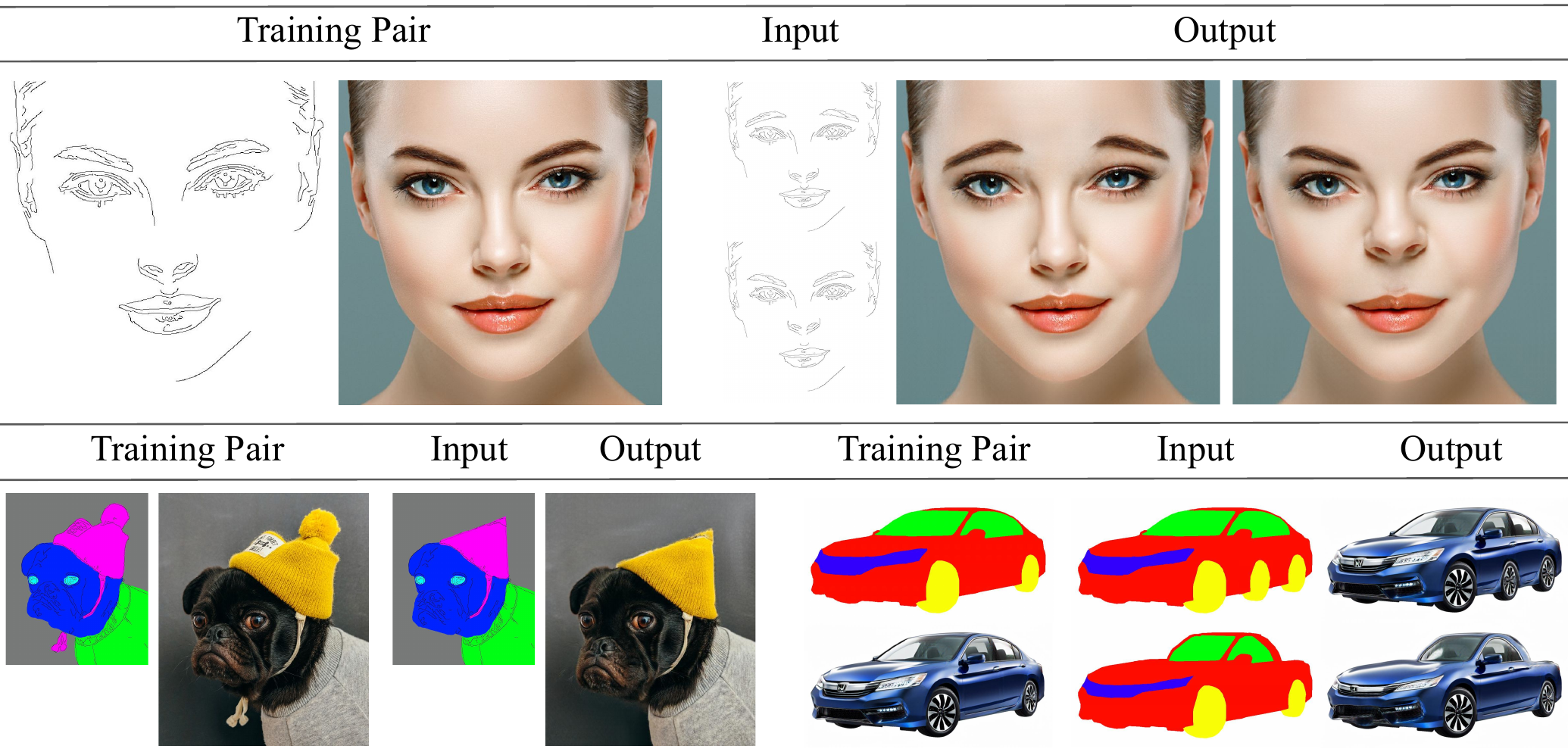}
\end{center}
\caption{\textit{Results produced by our model.} The model was trained on a single training pair, shown to the left of each sample. First row "face" output- (left) flipping eyebrows, (right) lifting nose. Second row "dog" output- changing shape of dog's hat, removing ribbon, and making face longer. Second row "car" output-  (top) adding wheel, (bottom) conversion to sports car.}
\label{fig:main_table}
\end{figure*}

\section{Introduction}
\label{sec:intro}
\begin{figure*}[h]
\begin{center}
\includegraphics[width=1\linewidth]{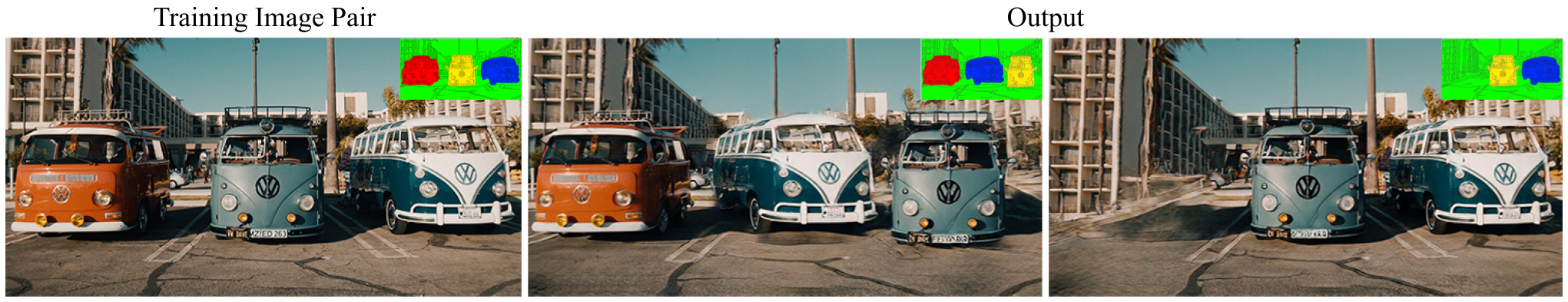}
\end{center}
\caption{\textit{Results on challenging manipulations.} Top right corners - primitive images. Left - original image used to train our model. Center- switching the positions between the two rightmost cars. Right- removing the leftmost car and inpainting the background. \textit{See the SM for many more results.}}
\label{fig:cars}
\end{figure*}

\blfootnote{*Equal contribution}
Deep neural networks have significantly boosted performance on image manipulation tasks for which large training datasets can be obtained, such as, mapping facial landmarks to facial images. In practice, however, there are many settings in which the image to be manipulated is unique, and a training set consisting of many similar input-output samples is unavailable. Moreover, in some cases using a large dataset might even lead to unwelcome outputs that do not preserve the specific characteristics of the desired image. Training generative models on just a single image, is an exciting recent research direction, which may hold the potential to extend the scope of neural-network-based image manipulation methods to unique images. In this paper, we introduce - DeepSIM, a simple-to-implement yet highly effective method for training deep conditional generative models from a single image pair.
Our method is capable of solving various image manipulation tasks including: (i) shape warping (Fig.~\ref{fig:main_table}) (ii) object rearrangement (Fig.~\ref{fig:cars}) (iii) object removal (Fig.~\ref{fig:cars}) (iv) object addition (Fig.~\ref{fig:main_table}) (v) creation of painted and photorealistic animated clips (Fig.~\ref{fig:animation_frames} and videos on our project page).

Given a single target image, first, a primitive representation is created for the training image. This can either be unsupervised (i.e. edge map, unsupervised segmentation), supervised (i.e. segmentation map, sketch, drawing), or a combination of both.
We use a standard conditional image mapping network to learn to map between the primitive representation and the image.  Once training is complete, a user can explicitly design and choose the changes they want to apply to the target image by manipulating the simple primitive (serving as a simpler manipulation domain). The modified primitive is fed to the network, which transforms it into the real image domain with the desired manipulation. This process is illustrated in Fig.~\ref{fig:main_im}.

Several papers have explored the topic of what and how much can be learned from a single image. Two recent seminal works SinGAN \cite{shaham2019singan} and InGAN \cite{inGAN} propose to extend this beyond the scope of texture synthesis \cite{BergmannJV17, jetchev2017texture,LiW16, zhou2018nonstationary}.
SinGAN tackles the problem of single image manipulation in an unconditional manner allowing unsupervised generation tasks. InGAN, on the other hand, proposes a conditional model for applying various geometric transformations to the image. Our paper extends this body of work by exploring the case of supervised image-to-image translation allowing the modification of specific image details such as the shape or location of image parts. We find that the augmentation strategy is key for making DeepSIM work effectively. Breaking from the standard practice in the image translation community of using a simple crop-and-flip augmentation, we found that using a thin-plate-spline (TPS) \cite{donato2002approximate} augmentation method is essential for training conditional generative models based on a single image-pair input. The success of TPS is due to its exploration of possible image manipulations, extending the training distribution to include the manipulated input. Our model successfully learns the internal statistics of the target image, allowing both professional and amateur designers to explore their ideas while preserving the semantic and geometric attributes of the target image and producing high fidelity results.

Our contributions in this paper:
\begin{itemize}
    \item A general purpose approach for training conditional generators supervised by merely a single image-pair. 
    \item Recognizing that image augmentation is key for this task, and the remarkable performance of thin-plate-spline (TPS) augmentation which was not previously used for single image manipulation.
    \item Achieving outstanding visual performance on a range of image manipulation applications. 
\end{itemize}

\begin{figure*}[t]
\centering
\includegraphics[width=0.8\linewidth]{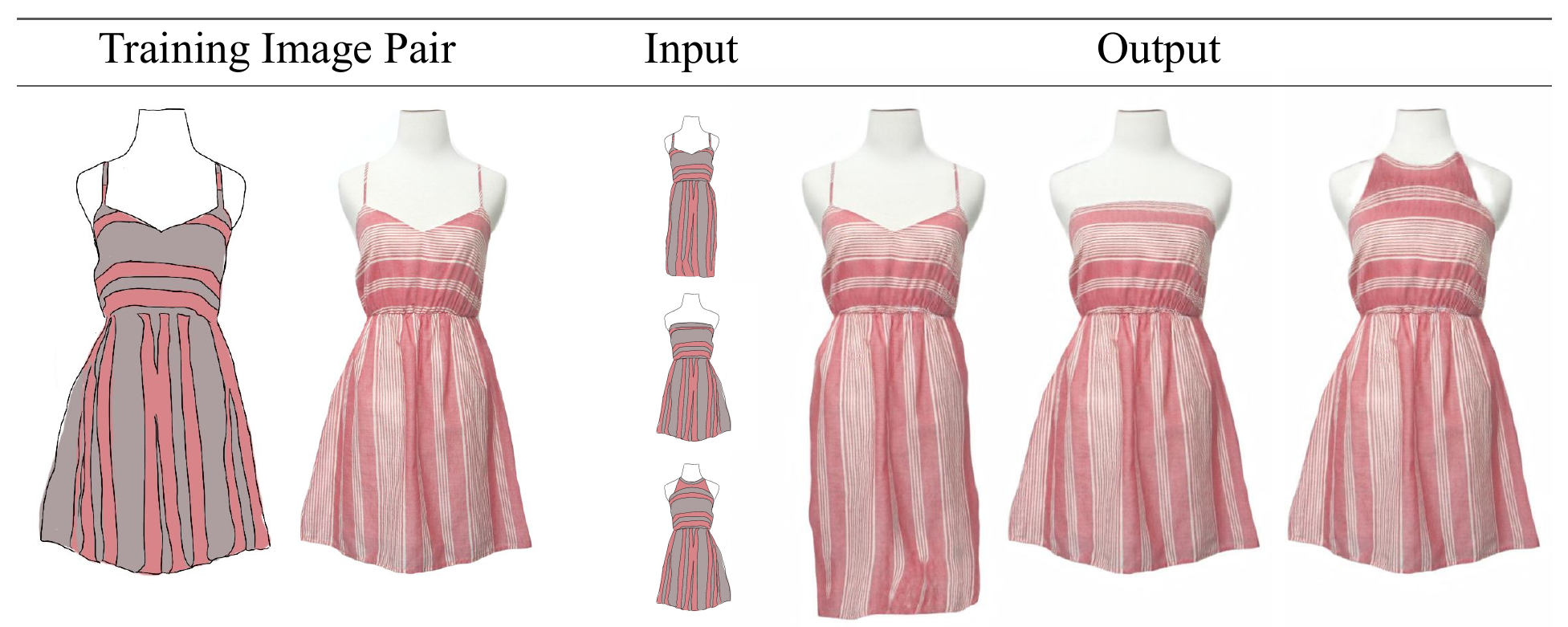}
 \caption{\textit{Fashion design examples.} On the left is the training image pair, in the middle are the manipulated primitives and on the right are the manipulated outputs- left to right: dress length, strapless, wrap around the neck.}
\label{tab:dress}
\end{figure*}

\begin{figure}[t]
\centering
\includegraphics[width=0.9\linewidth]{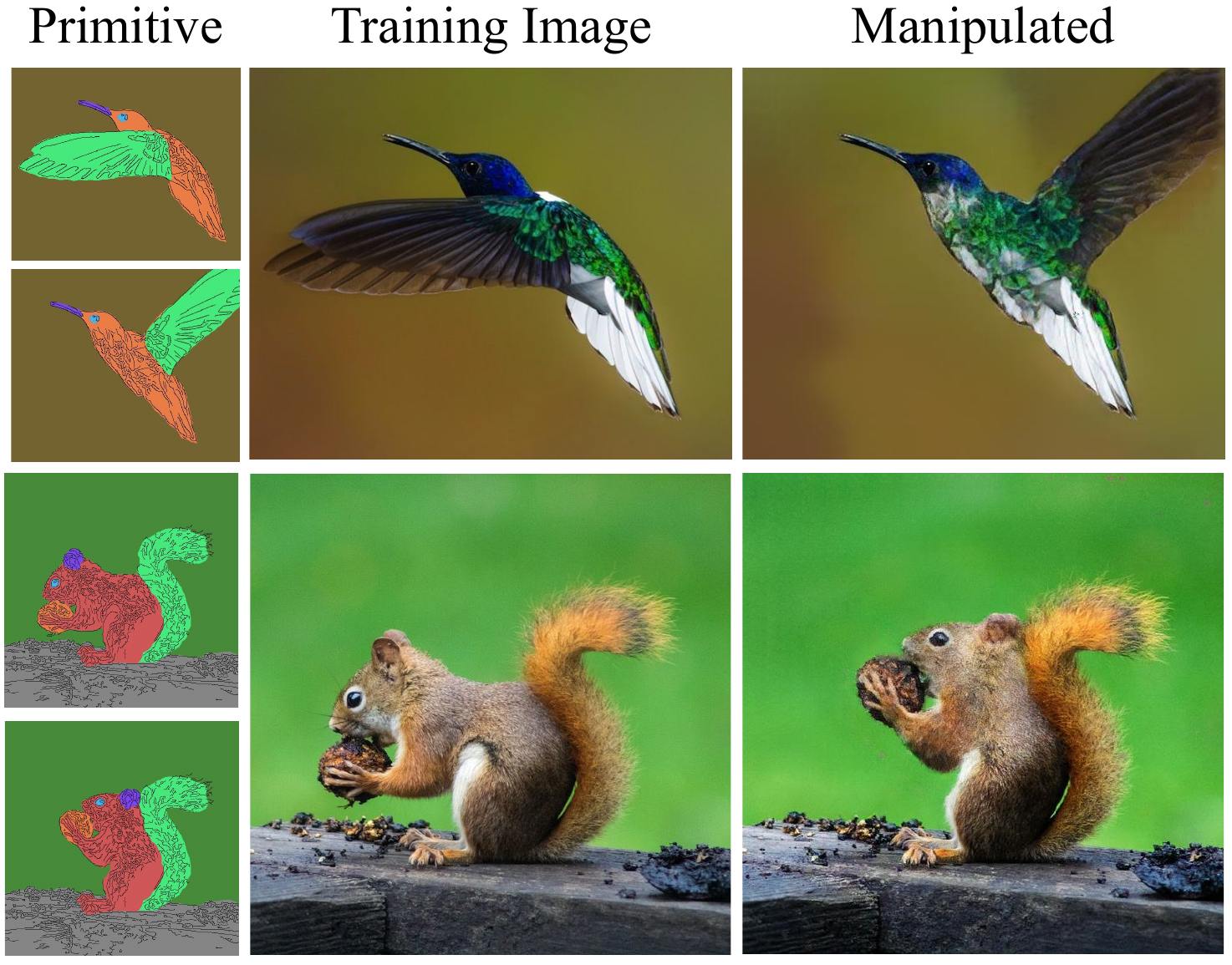}
 \caption{\textit{Natural looking manipulations.} Left: Image primitives, the top is the training primitive while the bottom is the manipulated one. Middle: training images. Right: manipulated outputs - changing the the orientation of the bird's wing, changing the posture of the squirrel.}
\label{tab:bird_squirrel_figure}
\end{figure}

\section{Related Work}
\label{sec:prev}

\textit{Classical image manipulation methods:} Image manipulation has attracted research for decades from the image processing, computational photography and graphics communities. It would not be possible to survey the scope of this corpus of work in this paper. We refer the reader to the book by \cite{szeliski2010computer} for an extensive survey, and to the Photoshop software for a practical collection of image processing methods. A few notable image manipulation techniques include: Poisson Image Editing \cite{perez2003poisson}, Seam Carving \cite{avidan2007seam}, PatchMatch \cite{barnes2009patchmatch}, ShiftMap \cite{pritch2009shift}, and Image Analogies \cite{analogies}. Spline based methods include: Field Morphing \cite{featue-morphing} and Image Warping by RDBF \cite{rdbf-warp}. Learning a high-resolution parametric function between a primitive image representation and a photo-realistic image was very challenging for pre-deep learning methods.

\textit{Deep conditional generative models:} Image-to-image translation maps images from a source domain to a target domain, while preserving the semantic and geometric content of the input images. Most image-to-image translation methods use Generative Adversarial Networks (GANs) \cite{goodfellow2014generative} that are used in two main scenarios: i) unsupervised image translation between domains \cite{CycleGAN2017, discogan, liu2017unsupervised, choi2017stargan} ii) serving as a perceptual image loss function \cite{pix2pix, pix2pixHD, spade, zhu2017toward}. Existing methods for image-to-image translation  require many labeled image pairs. Several methods \cite{chen2018sketchygan, dekel2017smart, zhu2018generative} are carefully designed for image manipulation, however they require large datasets which are mainly available for faces or interiors and cannot be applied to the long-tail of images.

\textit{Non-standard augmentations:} Conditional generation models typically use crop and flip augmentations. Classification models also use chromatic and noise augmentation. Recently, methods have been devised for learning augmentation for classification tasks e.g. AutoAugment \cite{cubuk2018autoaugment}. \cite{mounsaveng2019adversarial} learned warping fields for augmenting classification networks. Thin-plate-spline transformation have been used in the medical domain e.g. \cite{tang2019augmentation}, but they are used for training on large datasets rather than a single sample. \cite{zhao2019data} learned augmentations for training segmentation networks from a single annotated 3D medical scan (using a technique similar to \cite{kanazawa2016warpnet}) however they require a large unlabeled dataset of similar scans which is not available in our setting. TPS has also been used as a way of parametrizing warps for learning dense correspondences between images e.g. \cite{han2018viton} and \cite{lee2020reference}.

\textit{Learning from a single image: } Although most deep learning works use large datasets, seminal works showed that single image training is effective in some settings. \cite{asano2019critical} showed that a single image can be used to learn deep features.  Limited work has been done on training image generators from a single image - Deep Image Prior \cite{ulyanov2018deep}, retargeting \cite{inGAN} and super-resolution \cite{shocher2018zero}. Recently, the seminal work, SinGAN \cite{shaham2019singan}, presented a general approach for single unconditional image generative model training. However its ability for conditional manipulation is very limited. TuiGAN \cite{tuigan}, on the other hand, proposed a conditional unsupervised image-to-image method based on a single image pair. However, their method requires retraining the network for every new pair. Our method, on the other hand, uses a single aligned image pair for training a single generator that can be used for multiple manipulations without retraining, it is able to affect significantly more elaborate changes to images including to large objects in the scene.

\section{DeepSIM: Learning Conditional Generators from a Single Image}
\label{sec:method}

Our method learns a conditional generative adversarial network (cGAN) using just a single image pair consisting of the main image and its primitive representation. To account for the limited training set, we augment the data by using thin-plate-spline (TPS) warps on the training pair.
The proposed approach has several objectives: i) single image training ii) fidelity - the output should reflect the primitive representation iii) appearance - the output image should appear to come from the same distribution as the training image. We will next describe each component of our method:

\subsection{Model:}  Our model design follows standard practice for cGAN models (particularly Pix2PixHD \cite{pix2pixHD}). Let us denote our training image pair $(x,y)$ where $y \in \mathbb{R}^{d_x \times d_y \times 3}$ is the input image ($d_x$ and $d_y$ are the number of rows and columns) and $x \in \mathbb{R}^{d_x \times d_y \times d_p}$ is the corresponding image primitive ($d_p$ is the number of channels in the image primitive). We learn a generator network $G:\mathbb{R}^{d_x \times d_y \times d_p} \rightarrow \mathbb{R}^{d_x \times d_y \times 3}$, which learns to map input image primitive $x$ to the generated image $G(x)$. The fidelity of the result is measured using the VGG perceptual loss $\ell_{perc}:(\mathbb{R}^{d_x \times d_y \times 3},\mathbb{R}^{d_x \times d_y \times 3}) \rightarrow \mathbb{R} $ \cite{Johnson2016Perceptual} , which compares the differences between two images using a set of activations extracted from each image using a VGG network pre-trained on the ImageNet dataset (we follow the implementation in \cite{pix2pixHD}). We therefore write the reconstruction loss $\ell_{rec}$:
\begin{equation}
\label{eq:vgg_loss}
    \ell_{rec}(x, y; G) = \ell_{perc}(G(x), y)
\end{equation}

\textit{Conditional GAN loss:} Following standard practice, we add an adversarial loss which measures the ability of a discriminator to differentiate between the (primitive, generated image) pair $(x, G(x))$ and the (primitive, true image) pair $(x, y)$. The conditional discriminator $D: (\mathbb{R}^{d_x \times d_y \times d_p},\mathbb{R}^{d_x \times d_y \times 3}) \rightarrow [0,1] $ is implemented using a deep classifier which maps a pair of primitive and corresponding image into the probability of the two being a ground truth primitive-image pair. $D$ is trained adversarially against $G$. The loss of the discriminator ($\ell_{adv}$) is:

\begin{equation}
\begin{split}
\label{eq:discriminator_loss}
    \ell_{adv}(x, y; D, G) = \log({D(x,y)})\\ + \log({1-D(x,G(x))})
\end{split}
\end{equation}

The combined loss $\ell_{total}$ is the sum of the reconstruction and adversarial losses, weighted by a  constant $\alpha$:

\begin{equation}
\begin{split}
\label{eq:mapping_loss}
    \ell_{total}(x, y; D, G) = \ell_{rec}(x, y; G) \\ +  \alpha \cdot \ell_{adv}(x, y; D, G)
\end{split}
\end{equation}

\begin{figure}[t]
\centering
\begin{tabular}{@{\hskip2pt}c@{\hskip2pt}c}
\includegraphics[width=0.35\linewidth]{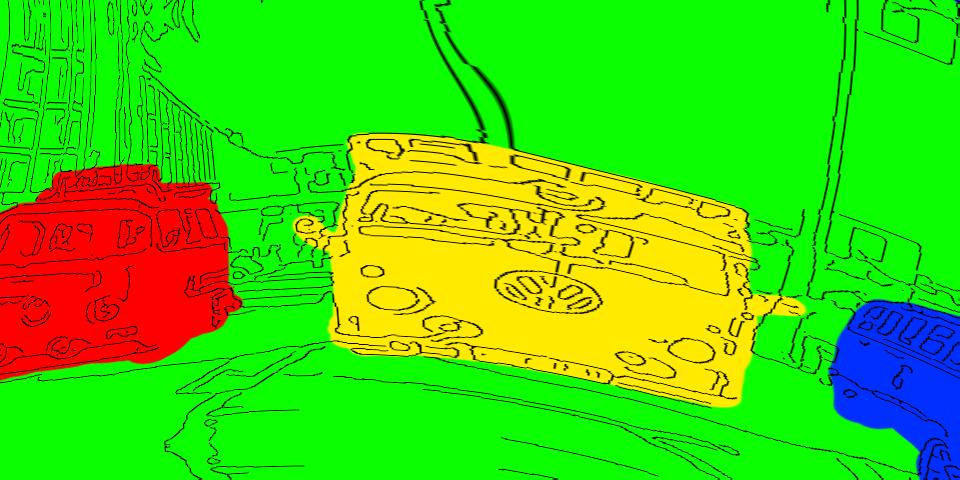} &
\includegraphics[width=0.35\linewidth]{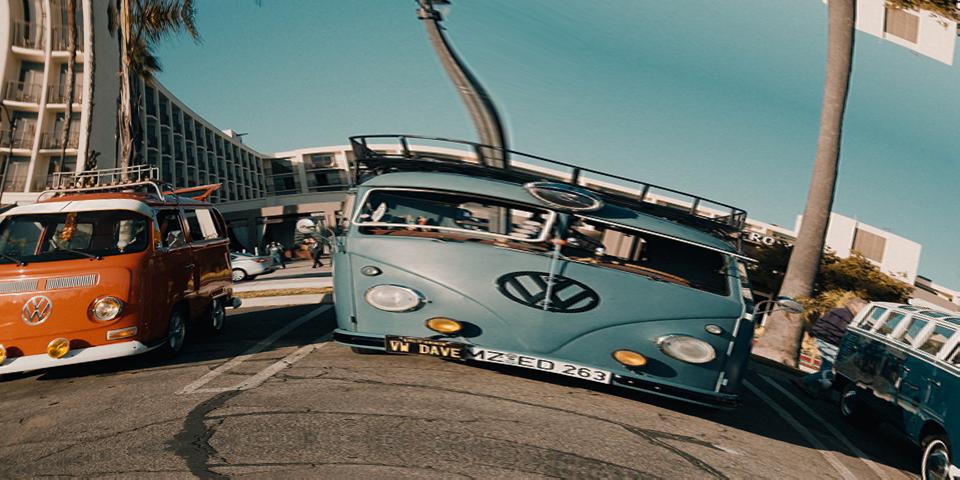} \\
\end{tabular}
 \caption{\textit{TPS Visualisation.} A random TPS warp of the primitive-image pair. Also see SM.}
\label{fig:tps}
\end{figure}

\begin{figure*}[t]
\begin{center}
\includegraphics[width=1.0\linewidth]{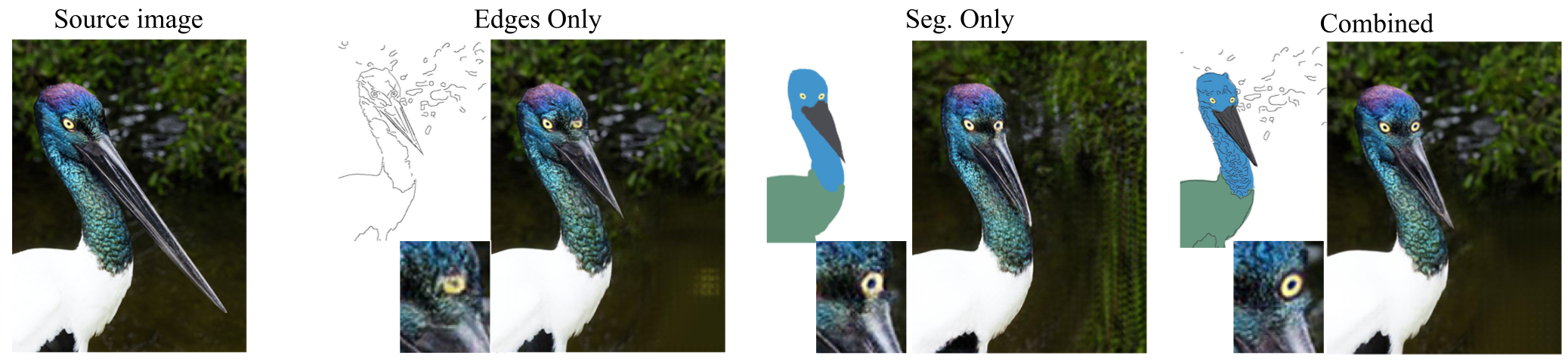}
\end{center}
\caption{\textit{Results on three different image primitives.} The leftmost column shows the source image, then each column demonstrate the result of our model when trained on the specified primitive. We manipulated the image primitives, adding a right eye, changing the point of view and shortening the beak. Our results are presented next to each manipulated primitive. The combined primitive performed best on high-level changes (e.g. the eye), and low-level changes (e.g. the background).}
\label{fig:bird}
\end{figure*}

\subsection{Augmentations:}
\label{subsec:aug}
When large datasets exist, finding the generator $G$ and conditional discriminator $D$ that optimize $\ell_{total}$ under the empirical data distribution can result in a strong generator $G$. However, as we only have a single image pair $(x, y)$, this formulation severely overfits. This has the negative consequence of $G$ not being able to generalize to new primitive inputs. In order to generalize to new primitive images, the size of the training dataset needs to be artificially increased so as to cover the range of expected primitives. Conditional generative models typically use simple crop-and-flip augmentations. We will later show (Sec.~\ref{sec:exp}) that this simple augmentation strategy however will not generalize to primitive images with non-trivial changes.

We incorporate the thin-plate-spline (TPS) as an additional augmentation in order to extend our single image dataset. For each TPS augmentation an equispaced $3\times3$ grid of control points $(i, j)$ is placed on the image, we then shift the control points by a random (uniformly distributed) number of pixels in the horizontal and vertical directions. This shift creates a \textit{non-smooth} warp which we denote by $t(i, j)$. To prevent the appearance of degenerate transformations in our training images, the shifting amount is restricted to at most $10\%$ of the minimum between the image width and height. We calculate the \textit{smooth} TPS interpolating function $f$ by minimizing:

\begin{equation}
\begin{split}
    \label{eq:tps_loss}
    \min_f\sum_{i, j} \|t(i, j) - f(i, j)\|^2 \\ + \lambda \int\int\Big(f_{xx}^2 + f_{yy}^2 + 2f_{xy}^2\Big) dx dy
\end{split}
\end{equation}
Where $f_{xx}, f_{xy}, f_{yy}$ denote the second order partial derivatives of $f$ which forms the smoothness measure, regularised by $\lambda$. The optimization over the warp $f$ can be performed very efficiently e.g. \cite{donato2002approximate}. We denote the distribution of random TPS that can be generated using the above procedure as $\Omega$. 
The above is illustrated in Fig. ~\ref{fig:tps}

\subsection{Optimization:}  During training, we sample random TPS warps. Each random warp $f \sim \Omega$ transforms both the input primitive $x$ and image $y$ to create a new training pair $(f(x), f(y))$ (where we denote $f(x)(i, j) = x(i', j')$ where $(i', j') = f(i, j)$). We optimize the generator and discriminator adversarially to minimize the expectation of the loss $\ell_{total}$ under the empirical distribution of random TPS warps:

\begin{equation}
D', G' = \min_G \max_D \mathbb{E}_{f \sim \Omega} \ell_{total}(f(x), f(y) ; D, G)
\end{equation}

We used the Pix2PixHD architecture with the official hyperparameters (except using $16000$ iterations).

\subsection{Primitive images:} To edit the image, we condition our generator on a representation of the image that we denote the image primitive. The required properties of the image primitive are: being able to precisely specify the required output image and the ease of manipulation by image editor. These two objectives are in conflict, although the most precise representation of the edited image is the edited image itself, this level of manipulation is very challenging to achieve by a human editor, in fact, simplifying this representation is the very motivation for this work. Two standard image primitives used by previous conditional generators are the edge representation of the image and the semantic instance/segmentation map of the image. Segmentation maps provide information on the high-level properties of the image, but give less guidance on the fine-details. Edge maps provide the opposite trade-off. To achieve the best of both worlds, we use the combination of the two primitive representations.
The advantages of the combined representation are shown in Sec.~\ref{sec:analysis}.
Our editing procedure is illustrated in the SM. 

\begin{figure*}[t]
\begin{center}
\includegraphics[width=1.0\linewidth]{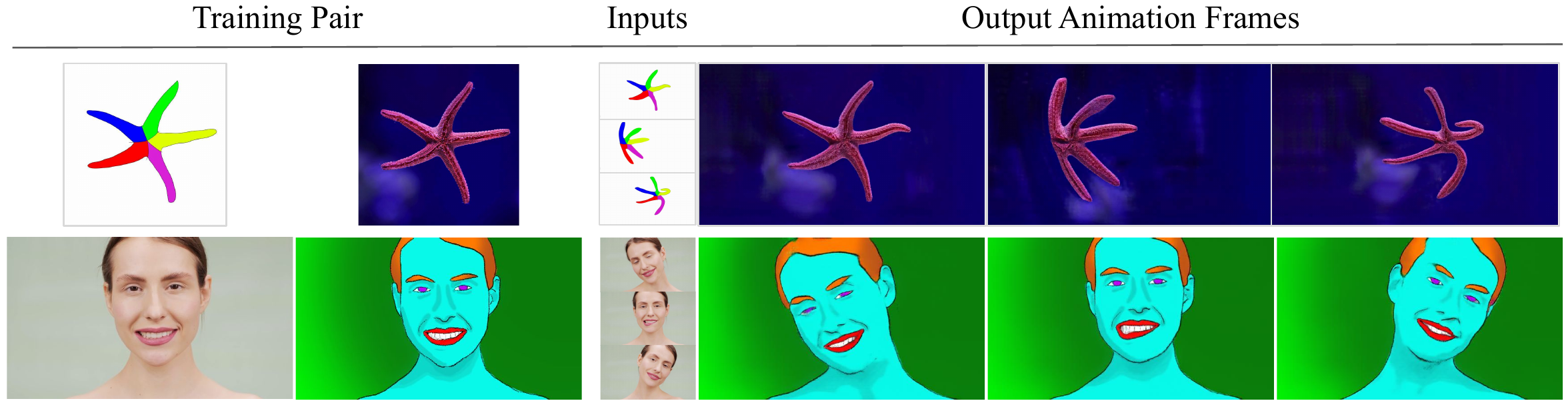}
\end{center}
\caption{\textit{Single Image Animation.} Top: translating an animation into a video clip, bottom- translating a video clip into a painted animation. Left: single training pairs, middle- subsequent frames, right: generated outputs. The video clips are available on our project page.}
\label{fig:animation_frames}
\end{figure*}

\begin{figure}[t]
\centering
\begin{tabular}{c|cc}
 Training Pair & Input & SinGAN \\
\includegraphics[width=0.25\linewidth]{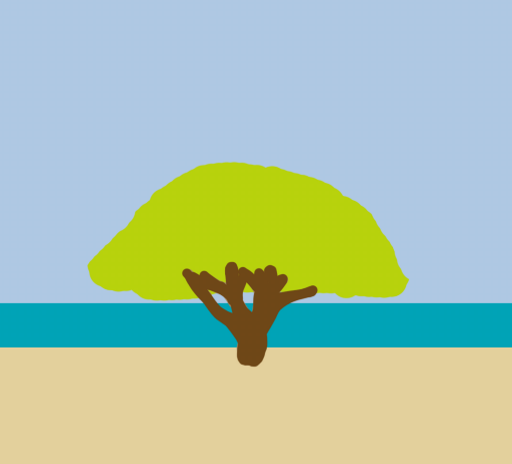} & \includegraphics[width=0.25\linewidth]{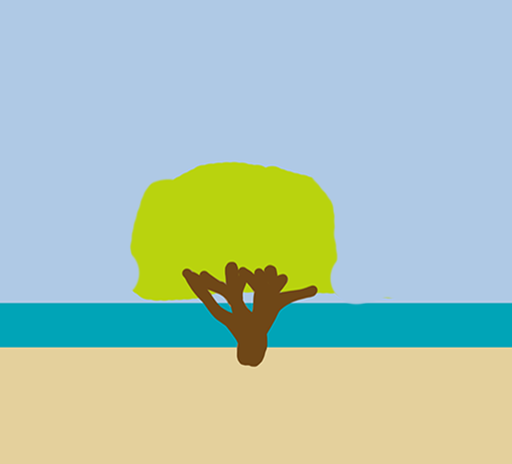} & \includegraphics[width=0.25\linewidth]{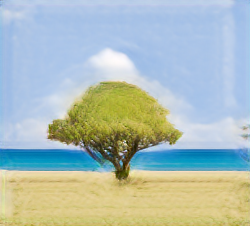} \\
  & TuiGAN & Ours \\
\includegraphics[width=0.25\linewidth]{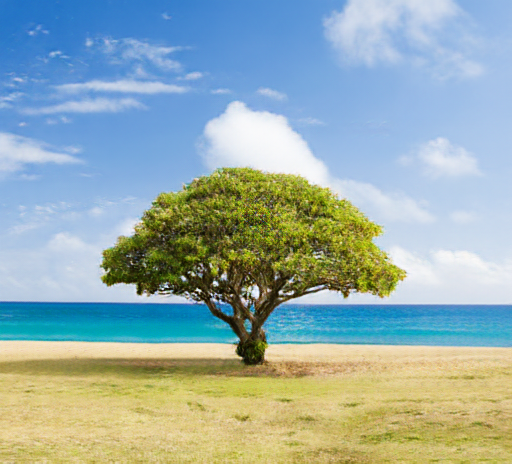} &
\includegraphics[width=0.25\linewidth]{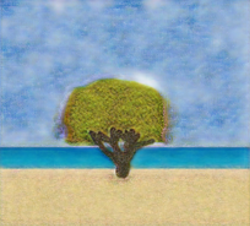} &
\includegraphics[width=0.25\linewidth]{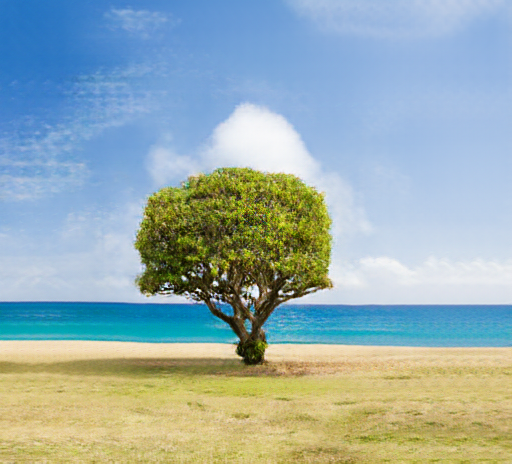}  \\

\end{tabular}
 \caption{\textit{Image manipulation comparison.} The leftmost column shows the training pair consisting of a painted image that was created manually and a target image. The manipulated image is given as input.
We can see that SinGAN preserves some details while failing to capture the shape, on the other hand, TuiGAN correctly captures the shape but does not preserve the details of the image. Our method is able to capture both the shape and the details of the manipulation with high fidelity.}
\label{tab:tree}
\end{figure}

\begin{figure}[t]
\centering
\footnotesize
\begin{tabular}{@{\hskip1pt}c@{\hskip1pt}c|@{\hskip1pt}c@{\hskip1pt}c@{\hskip1pt}c@{\hskip1pt}c} 

 \multicolumn{2}{c}{Training Image Pair} & Input & Pix2PixHD & BicycleGAN & Ours \\
 \multicolumn{2}{c}{}&&-MI&&\\

\includegraphics[width=0.15\linewidth]{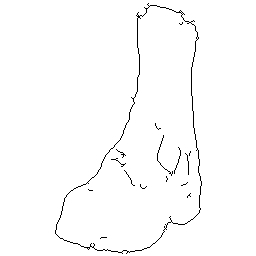} & 
\includegraphics[width=0.15\linewidth]{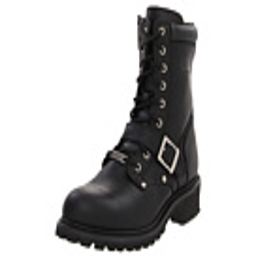} &
\includegraphics[width=0.15\linewidth]{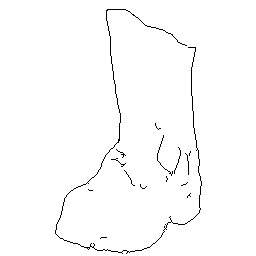} &
\includegraphics[width=0.15\linewidth]{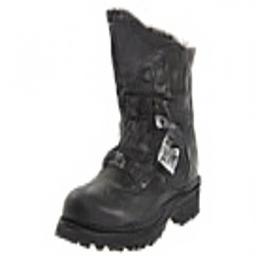} &
\includegraphics[width=0.15\linewidth]{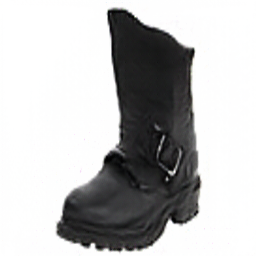} &
\includegraphics[width=0.15\linewidth]{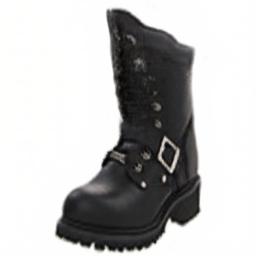}\\

\includegraphics[width=0.15\linewidth]{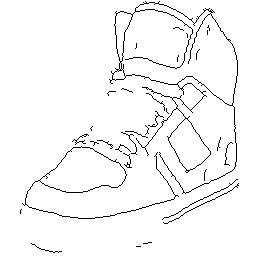} & 
\includegraphics[width=0.15\linewidth]{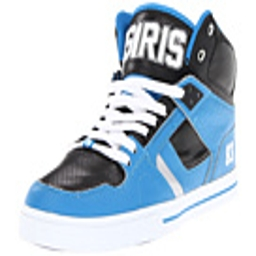} &
\includegraphics[width=0.15\linewidth]{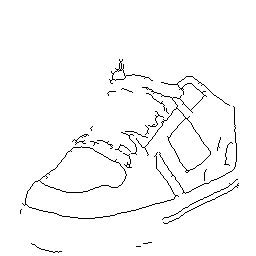} &
\includegraphics[width=0.15\linewidth]{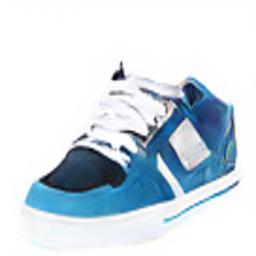} &
\includegraphics[width=0.15\linewidth]{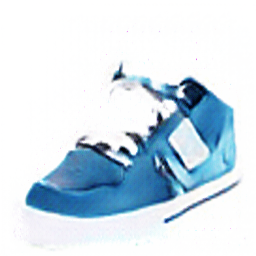} &
\includegraphics[width=0.15\linewidth]{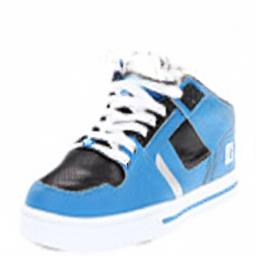}\\

\end{tabular}
 \caption{\textit{Edges-to-image comparison}. Columns $1,2$ show the training edges and images. Column $3$ shows the edges used as input at inference time. Pix2PixHD-MI cannot generate the correct shoe as there is not enough guidance. BicycleGAN has sufficient guidance but cannot reproduce the correct details. Our results are of high quality and fidelity.} \label{fig:shoes}
\end{figure}

\section{Experiments}
\label{sec:exp}

\subsection{Qualitative evaluation}

We present many results of our method in the main paper and SM. In Fig.~\ref{fig:main_table}, our method generates very high resolution results from single image training. In the top row we perform fine changes to the facial images from edge primitives e.g. raising the nose and flipping the eyebrows. In the second row, on the left, we used the combined primitive (edges and segmentation), we modify the dog's hat and made his face longer. On the right, we show complex shape transformations by using segmentation primitives. Our method added a third wheel to the car and converted its shape into a sports car. This shows the power of the segmentation primitive, enabling major changes to the shape using simple operations. See figures Fig.~\ref{tab:dress} and  Fig.~\ref{tab:bird_squirrel_figure} for more examples.

\begin{table*}[t]
\centering
\begin{center}
\begin{tabular}{lcccccccccc}
    \toprule
    Method     &  \multicolumn{2}{c}{S1}    &  \multicolumn{2}{c}{S2}  &  \multicolumn{2}{c}{S3}  &  \multicolumn{2}{c}{S4}  &  \multicolumn{2}{c}{S5} \\
    \cmidrule(lr){1-1} \cmidrule(lr){2-3} \cmidrule(lr){4-5} \cmidrule(lr){6-7} \cmidrule(lr){8-9} \cmidrule(lr){10-11}
    & L & S & L & S & L & S & L & S & L & S \\
     \midrule
     Pix2PixHD-SIA &              0.44 & 0.51       & 0.47   & 0.49     & 0.41     & 0.5       & 0.53      & 0.26     & 0.46   & 0.44\\
     Ours - no VGG &  0.14  & \textbf{0.05}   & 0.26 & \textbf{0.11 }      & 0.11      & 0.07  & 0.28  &   0.14        & 0.19    & 0.08  \\
      Ours &             \textbf{ 0.12}  &        0.07           & \textbf{0.21 }      & 0.12       & \textbf{ 0.1   }      & \textbf{0.04}  & \textbf{0.22 }  &   \textbf{0.12 }        &\textbf{ 0.14}                & \textbf{0.06 }  \\
            
    \bottomrule
\end{tabular}
\end{center}
 \caption{\textit{Quantitative comparison on LRS2 frames.} Results of Pix2PixHD-SIA (crop-and-flip) and our method (TPS) on $5$ LRS2 videos (both trained on a single pair). For each sequence left column: LPIPS, right column: SIFID.}
\label{tab:faces}
\end{table*}

\begin{figure*}[t]
\begin{center}
\includegraphics[width=1.0\linewidth]{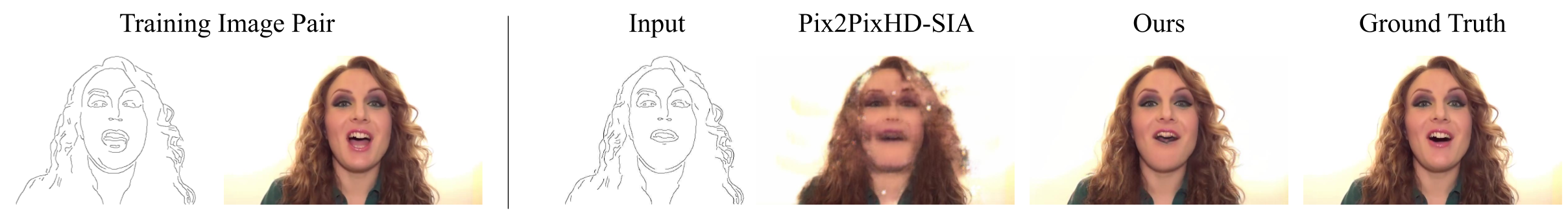}
\end{center}
\vspace{-2em}
\caption{\textit{Visually comparing the affect of TPS augmentations.} Our method with TPS outputs an image much more similar to the ground truth than just crop-and-flip augmentation (further results in SM).}
\label{fig:woman_vid}
\end{figure*}

In Fig.~\ref{tab:tree}, we compare the results of different single-image methods on a paint-to-image task. Our method was trained to map from a rough paint image to an image of a tree, while SinGAN and TuiGAN were trained using the authors' best practice. We can see that SinGAN outputs an image which is more similar to the paint than a photorealistic image and fails to capture the new shape of the tree. We note that although SinGAN allows for some conditional generation tasks, it is not its main objective, explaining the underwhelming results. TuiGAN on the other hand, does a better job in capturing the shape but fails to capture the fine details and texture. Our method is able to change the shape of the tree to correspond to the paint while keeping the appearance of the tree and background as in the training image. Differently from TuiGAN, we learn a single generator for all future manipulations of the primitive without the need to retrain for each manipulations.

In Fig.~\ref{fig:shoes}, we compare to two models that were trained on a large dataset. We can see that Pix2PixHD-MI (Pix2PixHD that was trained on the entire edge2shoes dataset, where "MI" is an acronym for "Multi Image") is unable to capture the correct identity of the shoes as there are multiple possibilities for the appearance of the shoe given the edge image. BicycleGAN is able to take as input both the edge map and guidance for the appearance (style) of the required shoe. Although it is able to capture the general colors of the required shoe, it is unable to capture the fine details of the shoes (e.g. shoe laces and buckles). This is a general disadvantage of training on large datasets, as a general mapping function becomes less specialized and therefore less accurate on individual images.

\textbf{Single Image Animation} the idea of generating short clip art  videos from only a single image was demonstrated in \cite{shaham2019singan} in an unsupervised fashion, we show that our model can be used to create an artistic short video clips in a supervised fashion from a single image-pair. This application allows to "breath life" in a single static image, by creating a short animated clip in the primitive domain, and feeding it frame-by-frame to the trained model to obtain a photorealistic animated clip. In contrast to SinGAN, which performs a random walk in the latent space, we allow for fine grained control over the animation "story".
In addition, our model can be used also in the opposite direction. That is, translating short video clips into painted animations based on a single frame and corresponding stylized image. This application may be useful for animators and designers. An example may be seen in Fig. \ref{fig:animation_frames}. We note that since our work does not focus on video generation, we do not have any temporal consistency optimization as was done by \cite{few-shot}. We strongly encourage the reader to view the videos on our project page.

\subsection{Quantitative evaluation}
\label{sec:quant}

As previous single image generators have mostly operated on unconditional generation, there are no established suitable evaluation benchmarks. We propose a new video-based benchmark for conditional single image generation spanning a range of scenes.  A single frame from each video is designated for training, where the network is trained to map the primitive image to the designated training frame. The trained network is then used to map from primitive to image for all the other video frames and compute the prediction error using LPIPS \cite{zhang2018unreasonable} and fidelity using SIFID \cite{shaham2019singan}. 

A visual evaluation on a frame from the LRS2 dataset can be seen in Fig.~~\ref{fig:woman_vid}. Our method is compared against Pix2PixHD-SIA, where "SIA" stands for "Single Image Augmented" e.g. a Pix2PixHD model that was trained on a single image using random crop-and-flip warps but not TPS. Our method significantly outperforms Pix2PixHD-SIA in fidelity and quality indicating that our TPS augmentation is critical for single image conditional generation. 
Quantitative evaluations on Cityscapes and LRS2 are provided in Tab.~\ref{tab:cityscapes} and Tab.~\ref{tab:faces}. We report LPIPS and SIFID for each of the $5$ LRS2 sequences and for the average of $16$ Cityscapes videos. Our method significantly outperformed Pix2PixHD-SIA in all comparisons.
More technical details may be found in the SM. SinGAN cannot perform this task and did not obtain meaningful results. While TuiGAN can in theory perform this task, it would require retraining a model for each frame which is impractical.  

\textbf{User Study} We conducted a user study, following the protocol of Pix2Pix and SinGAN. We sequentially presented 30 images: 10 real, 10 manipulated images, and 10 of side-by-side pairs of real and manipulated images. The participants were asked to classify each as “Real” or “Generated by AI”. In the case of pairs, we asked participants to determine if the ‘left’ or ‘right’ image was real. Each image was presented for $1$ second, as in previous protocols. The study consisted of $140$ participants. ($104$ males, $36$ females). The confusion rate on the unpaired images was $42.6\%$, while on the paired images it was $32.6\%$. This shows that our manipulated images are very realistic.

\begin{table}[t]
\centering
\begin{tabular}{lccc}
 
    \toprule
            Metric &  Pix2PixHD-SIA & \multicolumn{2}{c}{DeepSIM (Ours)} \\
            & Seg, Crop+Flip & Seg, TPS & Seg+Edge, TPS\\
             \midrule
            LPIPS    & 0.342 & 0.216 & \textbf{0.134} \\
            SIFID  & 0.292   & 0.127 & \textbf{0.104} \\
            
    \bottomrule
\end{tabular}

 \caption{Results for the Cityscapes dataset - we report the average over the 16 videos. The results show the importance of the TPS augmentation and the combined primitive.}
\label{tab:cityscapes}
\end{table}

\section{Analysis}
\label{sec:analysis}

\textit{Input primitives} As segmentations capture high-level aspects of the image while edge maps capture the low-level of the image better, we analyze the primitive that combines both. This choice is uncommon, e.g. Pix2PixHD proposed combining instance and semantic segmentation maps, however, this does not provide low-level details. 
Fig.~\ref{fig:bird} compares the three primitives. The edge representation is unable to capture the eye, presumably as it cannot capture its semantic meaning. The segmentation is unable to capture the details in the new background regions creating a smearing effect. The combined primitive is able to capture the eye as well as the low-level textures of the background region. In Fig.~\ref{fig:cars} we present more manipulation results using the combined primitive. In the center column, we switched the positions of rightmost cars. As the objects were not of the same size, some empty image regions were filled using small changes to the edges. A more extreme result can be seen in the rightmost column, the car on the left was removed, creating a large empty image region. By filling in the missing details using edges, our method was able to successfully complete the background (see SM for an ablation).

\textit{Runtime}: 
Our runtime is a function of the neural architecture and the number of iterations. When running all experiments on the same hardware (NVIDIA RTX-2080 Ti), a 256x256 image e.g. the "face" image (Fig.~\ref{fig:main_table}) takes SinGAN $72$ minutes to train, and $180$ minutes for TuiGAN while DeepSIM (ours) takes $49$ minutes. As was discussed previously, TuiGAN requires a new training process for each new manipulation whereas our DeepSIM does not.

\textit{Is the cGAN loss necessary?} We evaluated removing the cGAN loss, keeping just the VGG perceptual loss on the Cars image (see SM). For such high-res images the cGAN was a better perceptual loss. At lower resolutions, the VGG results were reasonable but still blurrier than the cGAN loss.

\begin{figure}[t]
\centering
\footnotesize
\begin{tabular}{@{\hskip1pt}c@{\hskip1pt}c@{\hskip1pt}c@{\hskip1pt}|@{\hskip1pt}c@{\hskip1pt}c@{\hskip1pt}|@{\hskip1pt}c@{\hskip1pt}c}

\rotatebox[ origin=c]{90}{Training}& \includegraphics[align=c, width=0.15\linewidth]{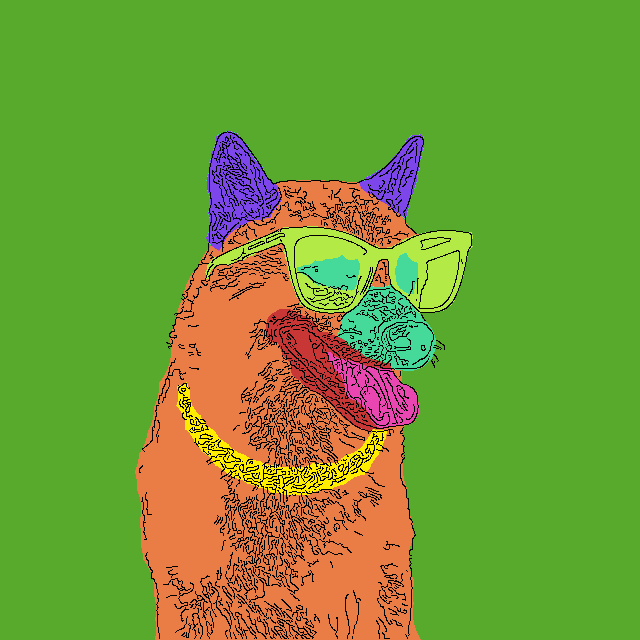} & 
\includegraphics[align=c, width=0.15\linewidth]{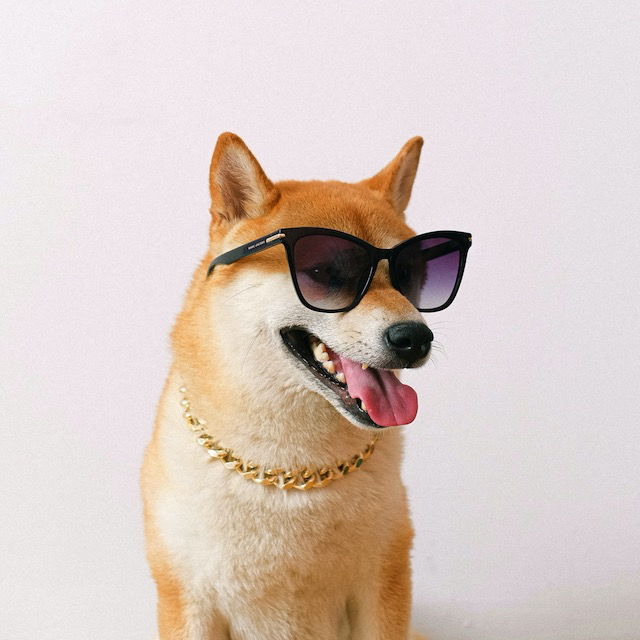} & \includegraphics[align=c, width=0.15\linewidth]{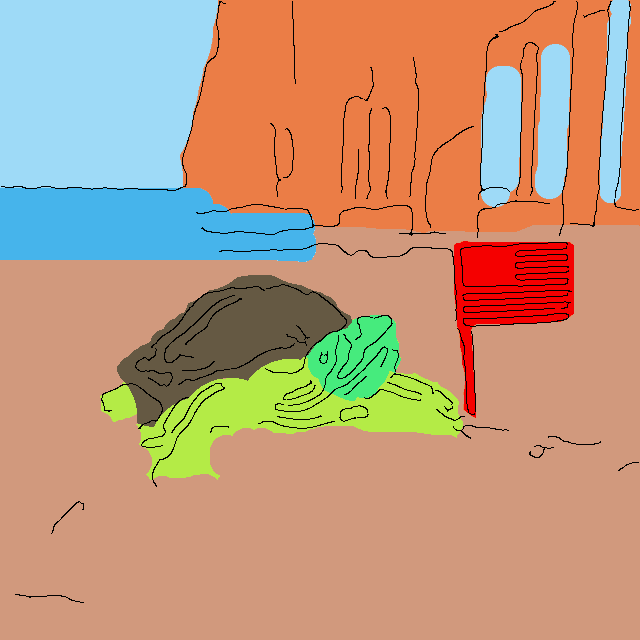} & 
\includegraphics[align=c, width=0.15\linewidth]{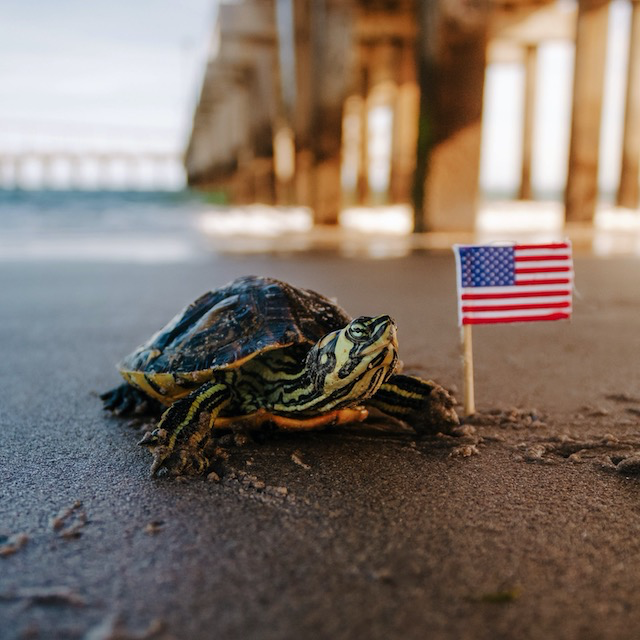} & 
\includegraphics[align=c, width=0.16\linewidth]{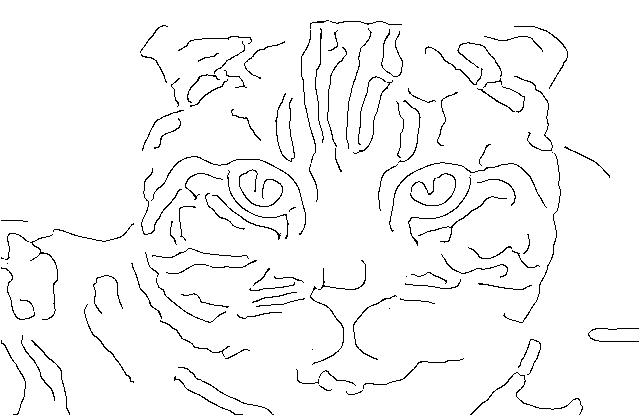} & 
\includegraphics[align=c, width=0.16\linewidth]{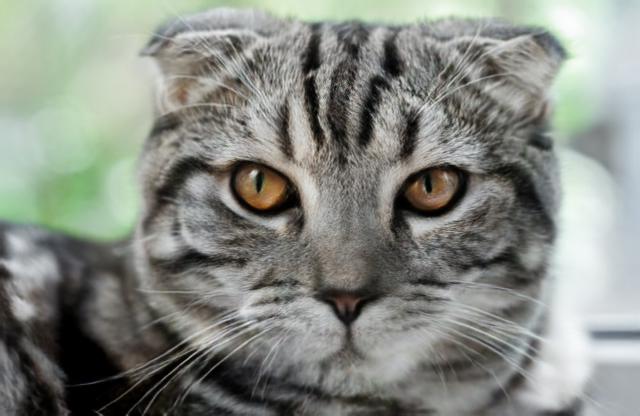}\\

\rotatebox[ origin=c]{90}{Output} &
\includegraphics[align=c, width=0.15\linewidth]{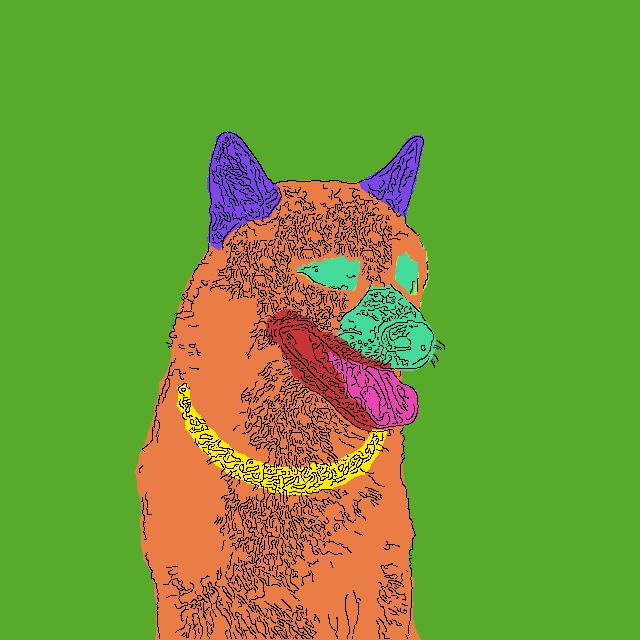} & 
\includegraphics[align=c, width=0.15\linewidth]{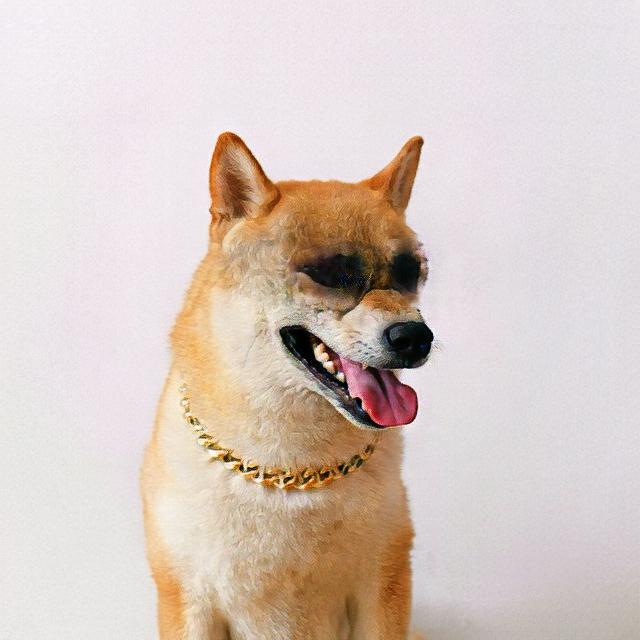} & 
\includegraphics[align=c, width=0.15\linewidth]{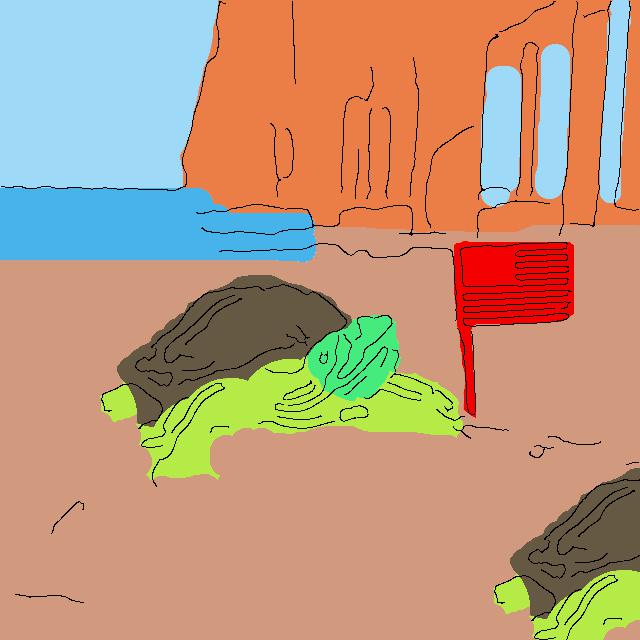} & 
\includegraphics[align=c, width=0.15\linewidth]{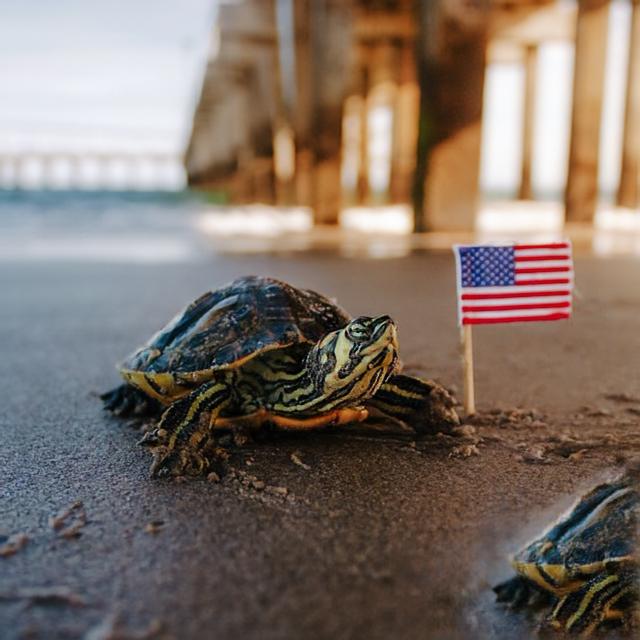} & 
\includegraphics[align=c, width=0.16\linewidth]{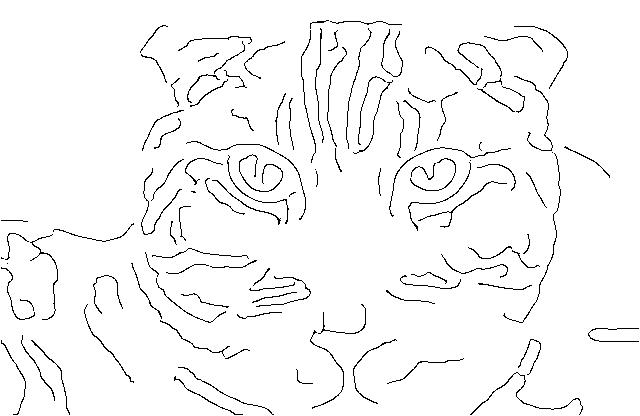} & 
\includegraphics[align=c, width=0.16\linewidth]{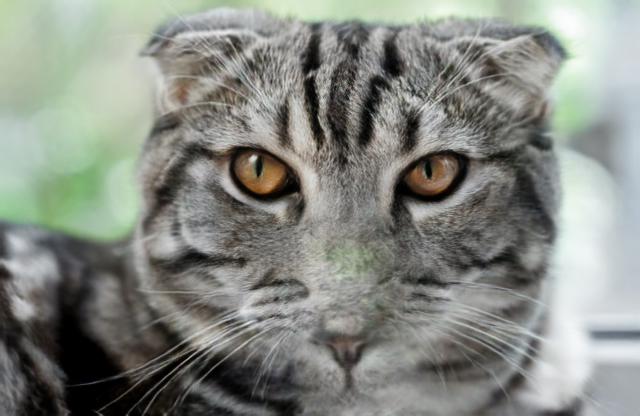} \\

\end{tabular}
 \caption{\textit{Failure modes.} Left: generating unseen objects - eyes of the dog. Center: background duplication - sea behind the turtle. Right: empty space interpolation - nose of the cat.} \label{fig:failure}
\end{figure}

\textit{Can methods trained on large datasets generalize to rare images?} We present examples where this is not the case. Fig.~\ref{fig:shoes} showed that BicycleGAN did not generalize as well as Pix2PixHD-MI for new (in-distribution) shoes. We show that in the more extreme case, where the image lies further from the source distribution used for training, current methods fail completely. See SM for further analysis. 

\textit{Augmentation in deep single image methods:} Although we are the first to propose single-image training for manipulation using extensive non-linear augmentations, we see SinGAN as implicitly being an augmentation-based \textit{unconditional} generation approach. In its first level it learns an unconditional low-res image generator, while latter stages can be seen as an upscaling network. Critically, it relies on a set of “augmented” input low-res images generated by the first stage GAN. Some other methods e.g. Deep Image Prior do not use any form of augmentation.

\textit{Failure modes:} We highlight three main failure modes of DeepSIM (Fig.~\ref{fig:failure}): i) generating unseen objects - when the manipulation requires generating objects unseen in training, the network can do so incorrectly. ii) background duplication - when adding an object onto new background regions, the network can erroneously copy some background regions that originally surrounded the object. iii) interpolation in empty regions - as no guidance is given in empty image regions, the network hallucinates details, sometimes incorrectly. See SM for further analysis.

\section{Conclusions}
\label{sec:conc}

We proposed a method for training conditional generators from a single training image based on TPS augmentations. Our method is able to perform complex image manipulation at high-resolution. Single image methods have significant potential, they preserve image fine-details to a level not typically achieved by previous methods trained on large datasets. One limitation of single-image methods (including ours) is the requirement for training a separate network for every image. Speeding up training of single-image generators is a promising direction for future work.

\textbf{Acknowledgements} We thank Jonathan Reich for creating the primitives and the animations examples and Prof. Shmuel Peleg for insightful comments and advise.

{\small
\bibliographystyle{ieee_fullname}
\bibliography{egbib}
}

\end{document}